\documentclass[10pt,letterpaper]{article}

\usepackage{ccn}
\usepackage{pslatex}
\usepackage{apacite}
\usepackage{graphicx}

\title{The Algonauts Project: A Platform for Communication between \\ the Sciences of Biological and Artificial Intelligence}
 
\author{{\large \bf Radoslaw Martin Cichy$^{1,*}$ (radoslaw.cichy@fu-berlin.de), Gemma Roig$^{2}$ (gemmar@mit.edu), Alex}\\
{\large \bf Andonian$^{3}$ (aandonia@mit.edu), Kshitij Dwivedi$^{2}$ (kshitij\_dwivedi@mymail.sutd.edu.sg), Benjamin }\\
{\large \bf Lahner$^{3}$ (blahner@mit.edu), Alex Lascelles$^{3}$ (alexlasc@mit.edu), Yalda Mohsenzadeh$^{3}$}\\
{\large \bf  (yalda@mit.edu), Kandan Ramakrishnan$^{3}$ (krama@mit.edu), Aude Oliva$^{3}$ (oliva@mit.edu)}\\
$^{1}$Department of Education and Psychology, Freie Universit{\"a}t Berlin, Berlin, Germany\\
$^{2}$Information Systems Technology and Design, Singapore University Technology and Design, Singapore\\
$^{3}$Computer Science and Artificial Intelligence Laboratory, MIT, Cambridge, USA\\
$^*$ corresponding author}

\begin{document}

\maketitle

\section{Abstract}
{
\bf
In the last decade, artificial intelligence (AI) models inspired by the brain have made unprecedented progress in performing real-world perceptual tasks like object classification and speech recognition. Recently, researchers of natural intelligence have begun using those AI models to explore how the brain performs such tasks. These developments suggest that future progress will benefit from increased interaction between disciplines. Here we introduce the Algonauts Project as a structured and quantitative communication channel for interdisciplinary interaction between natural and artificial intelligence researchers. The project's core is an open challenge with a quantitative benchmark whose goal is to account for brain data through computational models. This project has the potential to provide better models of natural intelligence and to gather findings that advance AI. The 2019 Algonauts Project focuses on benchmarking computational models predicting human brain activity when people look at pictures of objects. The 2019 edition of the Algonauts Project is available online:
\url{http://algonauts.csail.mit.edu/}.
}
\begin{quote}
\small
\textbf{Keywords:} 
human neuroscience; vision; object recognition; prediction; challenge; competition; benchmark
\end{quote}

\section{Introduction}

The quest to understand the nature of human intelligence and engineer advanced forms of artificial intelligence (AI) are increasingly intertwined \cite{Hassabis17, Kriegeskorte15, Yamins16}. To explain human intelligence, we require computational models that can handle the complexity of real-world tasks. To engineer artificial intelligence, biological systems can provide inspiration and guidance of how to solve the task efficiently.

With this algorithmic exploration paradigm for explaining the brain, it is becoming essential to have standardized benchmarks for comparing how well different algorithms account for neural data. Open challenges are a particular form of standardized benchmark that foster fast-paced advance in a collaborative and transparent manner.

Open challenges have helped science to thrive in many times and fields. As early as 1900, Hilbert proposed 23 problems as challenges in mathematics to be solved. More recently, benchmarks for open competition have emerged in other disciplines such as robotics (e.g. the DARPA robotics challenge) and computer science on a diverse sets of topics including visual recognition ~\cite{Everingham15,ILSVRC15,PlacesChallenge}), reasoning~\cite{CLEVR} and natural language understanding~\cite{GLUE}. Those challenges are well accepted in the their scientific communities and suggest standardized benchmarks as fruitful platforms for collaboration.

Inspired by these approaches, we propose a challenge platform with standardized benchmarks for the artificial and biological sciences.  At the core of the platform is an open competition with the goal of accounting for brain activity through computational models and algorithms. We coin the platform the Algonauts Project. Inspired by the astronauts (i.e. sailors of the stars) who launched into space to explore a new frontier, the algonauts (i.e. sailors of algorithms)  set out to relate brains and computer algorithms in an exploratory way. 

We believe that the Algonauts Project will facilitate the interaction between biological and artificial intelligence researchers, allowing the communities to exchange ideas and advance both fields rapidly and in a transparent way.

\section{The 2019 Edition of the Algonauts Project: Explaining the Human Visual Brain}

The 2019 edition is the first edition of the Algonauts Project's challenge and workshop. It is titled "Explaining the Human Visual Brain", and its specific target is to determine which computational model best accounts for human visual brain activity.

We focus on visual object recognition as it is an essential cognitive capacity of systems embedded in the real world. Visual object recognition has long fascinated neuroscientists and computer scientists alike, and it is here that the recent advances in AI and their adoption into neurosciences have taken place most prominently. Currently, particular deep neural networks trained with the engineering goal to recognize objects in images do best in accounting for brain activity during visual object recognition \cite{Schrimpf18,Bashivan19}. However, a large portion of the signal measured in the brain remains unexplained. This is so because we do not have models that capture the mechanisms of the human brain well enough. Thus, what is needed are advances in computational modelling to better explain brain activity.

\subsubsection{Related challenges in neuroscience.}

The 2019 edition "Explaining the Human Visual Brain" relates to initiatives such as the “The neural prediction challenge” (\url{http://neuralprediction.berkeley.edu/}) and “brain-score” (\url{http://www.brain-score.org/}) \cite{Schrimpf18} that provide benchmarks and leaderboards. The Algonauts Project emphasizes human brain data, and an automated submission procedure with immediate assessment. It  couples neural prediction benchmarks to a challenge limited in time, and adds educational and collaborative components through the accompanying workshop.

\section{Materials and Methods}

The target of the 2019 challenge is to account for activity in the human visual brain responsible for object recognition. This is the so-called ventral visual stream \cite{Grill04}, a hierarchically ordered set of brain regions in which neural activity unfolds across regions in space and time when human beings see an object. It starts with early visual cortex (EVC) and continues in inferior temporal (IT) cortex. Neurons in EVC respond preferentially to simple visual features such as oriented edges, whereas neurons in IT respond to more complex and larger features such as object parts. Consistent with their position in the processing hierarchy, neurons in EVC have been found to respond to visual stimulation earlier in time than neurons in IT. Stages of brain processing can thus be identified both in space (different regions) and in time (early and late). Correspondingly we have two challenge tracks.

\noindent{\bf Track 1} aims to account for brain data in space, providing data from the start and later point of the ventral visual stream: early visual cortex (EVC) and inferior temporal cortex (IT), respectively (Fig.~\ref{fig1}a). We provide brain data measured with functional magnetic resonance imaging (fMRI\footnote{For more details on MEG and fMRI see \url{http://algonauts.csail.mit.edu/fmri\_and\_meg.html}}), a technique with high spatial resolution (millimeters) that measures blood flow changes associated with neural activity.

\noindent{\bf Track 2} aims to account for brain data in time, providing data recorded early and late in visual processing (Fig.~\ref{fig1}a). For this we provide brain data measured with magnetoencephalography (MEG) at time points identified to correspond to processing in EVC and IT. MEG is a technique with very high temporal resolution (milliseconds) that measures the magnetic fields accompanying electrical activity in the brain.

\subsubsection{Comparison metric from brain activity and models to challenge score.}

\begin{figure}[t!]
\begin{center}
\includegraphics[width=\linewidth]{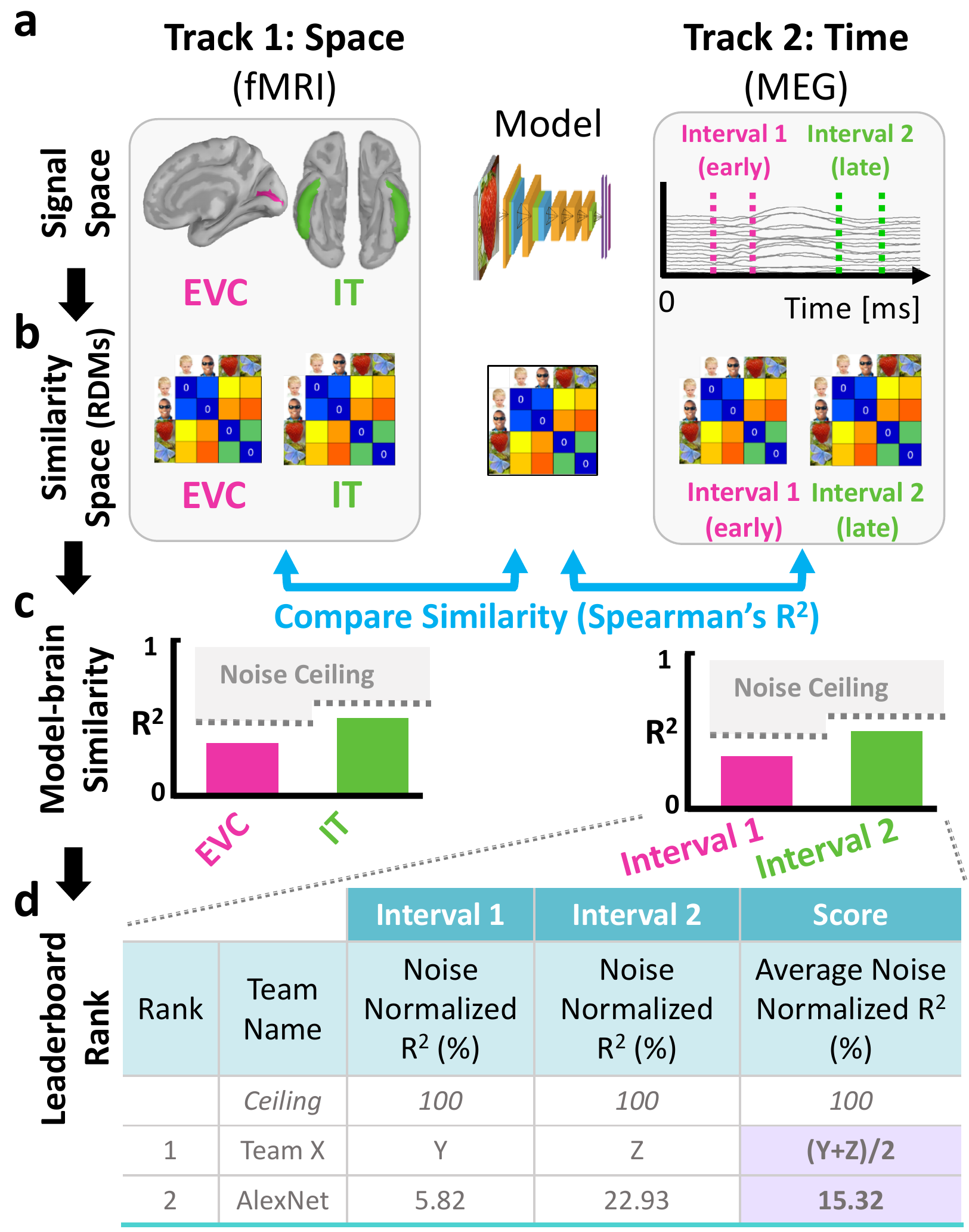}
\end{center}
\caption{\emph{Procedure of the Algonauts 2019 edition challenge.} {\bf a)} In two tracks, the goal is to account for human brain activity measured during object perception in space and time. {\bf b)} RSA makes models and brain activity comparable, yielding {\bf c)} percent variance explained relative to the noisiness of the data. {\bf d)} Models are ranked in a leaderboard (i.e. for Track 2).} 
\label{fig1}
\end{figure}

Comparing human brains and models is challenging because of the numerous differences between them (e.g. in-silico vs. biological, number of units). Different approaches have been proposed \cite{Diedrichsen17, Wu06}, and here we make use of a technique called representational similarity analysis (RSA) \cite{Kriegeskorte08, Kriegeskorte13}. RSA has low computational demands and is straightforward to implement. The idea behind RSA is that models and brains are similar if they treat the same images as similar (or equivalently dissimilar). RSA is a two-step procedure. In a first step (Fig. \ref{fig1}b), we abstract from the incommensurate signal spaces into similarity space by calculating pairwise dissimilarities between signals for all conditions (images) and order them in so-called representational dissimilarity matrices (RDMs) indexed in rows and columns by the conditions compared. RDMs for the different signal spaces have the same dimensions and are directly comparable. We relate RDMs in a second step (Fig. \ref{fig1}c) by calculating their similarity (Spearman’ R). Finally, we square the result to R$^2$ to indicate the amount of variance explained, and display results in the leaderboard (Fig. \ref{fig1}d).

\subsubsection{Noise ceiling.} 

The noise ceiling is the expected RDM correlation achieved by the (unknown) ideal model, given the noise in the data. The noise ceiling is computed by the assumption that the subject-averaged RDM is the best estimate of the ideal model RDM, i.e. by averaging the correlation of each subject's RDM with the subject-averaged RDM. We use the noise ceiling to normalize R$^2$ values to noise-normalized variance explained. Thus, any model can explain from 0 to 100\% of the explainable variance.

\begin{figure}[t!]
\begin{center}
\includegraphics[width=\linewidth]{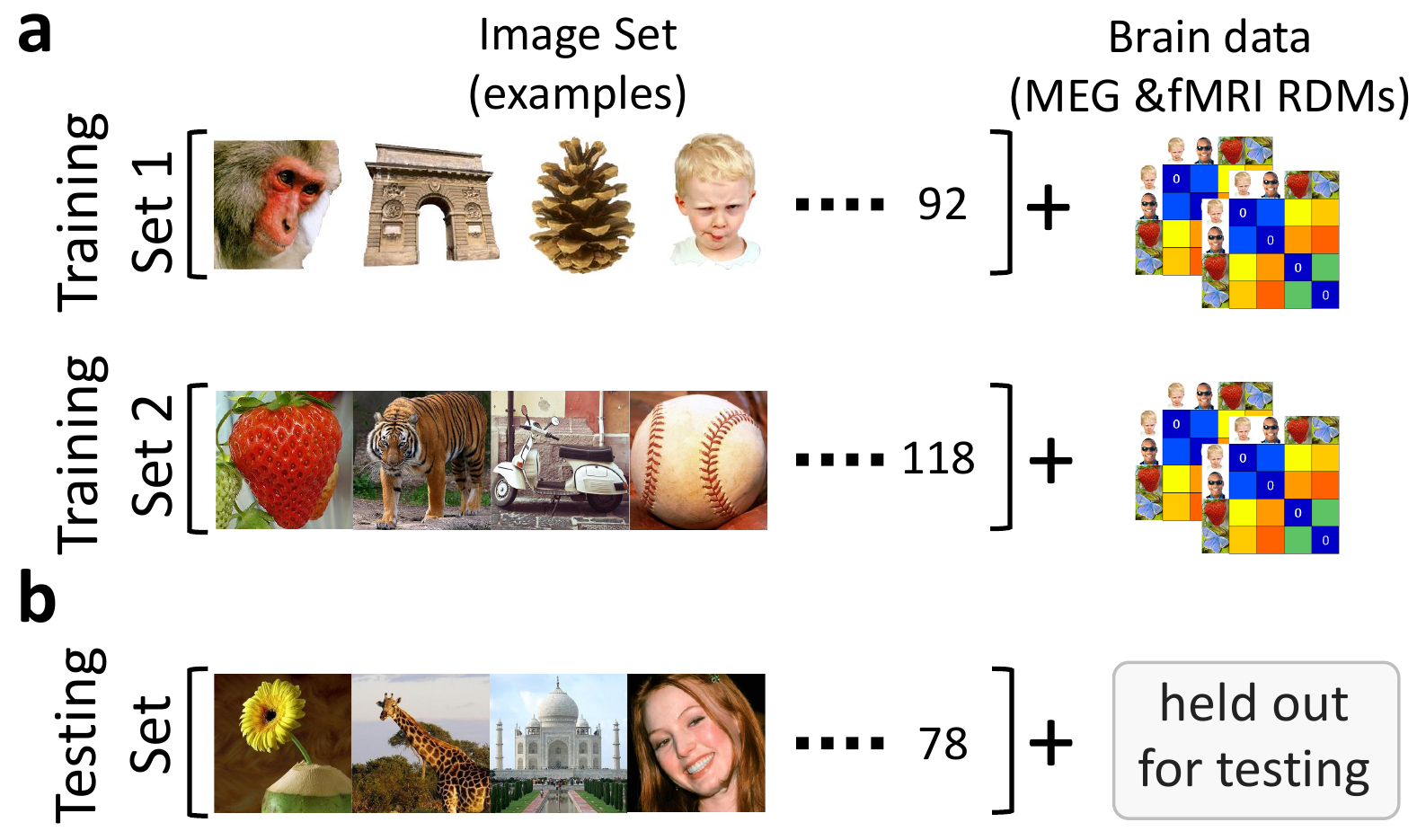}
\end{center}
\caption{\emph{Training and Testing Material.} {\bf a)} There are two sets of training data, each consisting of an image set and brain activity in RDM format (for fMRI and MEG). Training set 1 has 92 silhouette object images, and training set 2 has 118 object images with natural backgrounds. {\bf b)} Testing data consists of 78 images of objects on natural backgrounds. Associated brain data is held back and used to evaluate models online for the leaderboard.} 
\label{fig2}
\end{figure}

\subsubsection{Training Data.}

Participants can submit their models out of the box to determine how well they predict brain activity in each track. We also provide training data that can help optimizing models for predicting brain data. We provide two sets of training data published previously (Fig.~\ref{fig2}a) \cite{Cichy14, Cichy16}. Each set consists of a set of images (92 silhouette object images and 118 images of objects on natural background), and brain data recorded with fMRI (EVC and IT) and MEG (early and late in time) in response to viewing those images (by 15 participants). Participants differ across training sets but are the same across imaging modalities (MEG and fMRI).

\subsubsection{Testing Data and Procedure.} 

The testing set consists of 78 images and the respective brain activity recorded with fMRI and MEG (Fig. \ref{fig2}b). Participants in the challenge receive only the images, and the brain data is held back. On the basis of the image test set participants calculate model RDMs as predictions of human brain activity. Participants submit the RDMs which are compared against the held-out brain data using RSA as described above. This results in a challenge score and determines the relative place in the leaderboard.

\subsubsection{Rules.} 

To encourage broad participation the challenge consists of a simple submission process. Participants can use any model trained on any type of data, however we explicitly forbid the use of human brain responses to the test image set. We request participants to submit a short report to a preprint server describing their final submitted model.

\subsubsection{Development Kit.}

The development kit contains the aforementioned training and testing data. In addition, we provide example extraction code (matlab and python) to extract activation values from models into RDMs and evaluation code that compares model RDMs with brain RDMs, calculating the noise-normalized score for a model.

\subsubsection{Baseline Model.} 

Deep neural networks trained on object classification are currently the model class best performing in predicting visual brain activity. We used  \textit{AlexNet }\cite{Krizhevsky12} as an example often used in neuroscientific studies as baseline model. \textit{AlexNet} is a feedforward deep neural network, trained on object categorization, with 5 convolutional and 3 fully connected layers. In Track 1 (fMRI), \textit{AlexNet} accounts for 6.58\% (layer 2) and 8.22\% (layer 8) of  noise-normalized variance in EVC and IT. In track 2 (MEG), it accounts for 5.82\% (layer 2) and 22.93\% (layer 4) noise-normalized variance in early and late visual processing.

\section{Discussion}

\subsubsection{Challenges as scientific instruments in cognitive science.}

Open challenges at the intersection of natural and artificial intelligence sciences hold promise for both sides. The natural intelligence sciences, in particular neuroscience and psychology, might benefit in two ways. For one, open challenges provide the incentive structure to promote and ensure transparency and openness. These are values recognized to promote replicability of results \cite{Nosek15, Poldrack17}. Second, challenges provide a clear and quantitatively concise metric for success. They can thus play an important role in guiding research by differentiating between theories: predictive success is a necessary property of a good explanatory model \cite{Kriegeskorte15}.
The sciences creating artificial intelligence in turn might benefit, too, in several ways. Biological systems can provide insight into how a cognitive problem might be solved mechanistically. More specifically, neuroscience can provide constraints on the infinite number of free parameters when engineering a model from scratch. 

\subsubsection{Prediction vs. explanation.}

Challenges like the Algonauts Project provide one measure of success: predictive power. Having an artifact that even perfectly predicts a phenomenon does not by itself explain the phenomenon.  However, prediction and explanation are related goals \cite{Cichy19}. For one, successful explanations ultimately must also provide successful predictions \cite{Breiman01,Yarkoni17}. Second, the ordering of models on a challenge benchmark can help scientist to concentrate future research efforts in creating explanations based on the most successful models. Further, bringing success rate in connection with the models' properties can reveal what it is about those models that is responsible for the success. It can thus generate hypotheses and guide the next engineering steps. 

\subsubsection{Limitations of the current approach.} 

Constitutive for a challenge are the choice of a particular data set and analysis steps. We readily assert that we could have structured the challenge differently (e.g. which data to provide, in which format, how to relate brain data and models). The choices we made were motivated by providing a low threshold to participation and a low computational load. Future challenges that make use of other data sets (e.g. large-scale) will invite a different type of data format and analytic treatment. We will invite an open discussion on those issues during the workshop.

\subsubsection{The future of the project.} 

We hope that the 2019 edition of the Algonauts Project will inspire other researchers to initiate open challenges and collaborate with the Algonauts Project. We see potential in tackling problems that become increasingly interesting to both natural and artificial intelligence communities. In the context of perception, future challenges might put the focus on action recognition or involve other sensory modalities such as audition or the tactile sense, or focus on other cognitive functions such as learning and memory.

\vspace*{-0.12cm}
\section{Acknowledgments}

This research was funded by DFG (CI-241/1-1 CI-241/1-3) and an ERC grant (ERC-2018-StG 803370) to R.M.C; NSF award (1532591) in Neural and Cognitive Systems and the Vannevar Bush Faculty Fellowship program funded by the ONR (N00014-16-1-3116) to A.O. We thank our sponsors: the MIT Quest for Intelligence and the MIT-IBM Watson AI Lab. 

\vspace*{-0.12cm}
\bibliographystyle{apacite}

\setlength{\bibleftmargin}{.125in}
\setlength{\bibindent}{-\bibleftmargin}

\bibliography{ccn_style}

\end{document}